\pdfoutput=1

\documentclass[11pt]{article}

\usepackage{authblk}
\usepackage[review]{naacl2021}

\usepackage{times}
\usepackage{latexsym}
\usepackage{graphicx}
\usepackage{microtype}
\usepackage{multirow}
\usepackage{booktabs}
\usepackage{bm}
\usepackage[T1]{fontenc}

\usepackage[utf8]{inputenc}

\usepackage{microtype}

\title{
DiSCoL: Toward Engaging \textit{Di}alogue \textit{S}ystems through \textit{Co}nversational \textit{L}ine Guided Response Generation}

\author[1]{\textbf{Sarik Ghazarian}}
\author[1]{\textbf{Zixi Liu}}
\author[2]{\textbf{Tuhin Chakrabarty}} 
\author[1]{\textbf{Xuezhe Ma}}
\author[1]{\textbf{Aram Galstyan}}
\author[1,3]{\textbf{Nanyun Peng}}
\affil[1]{Information Sciences Institute, University of Southern California}
\affil[2]{Department of Computer Science, Columbia University  }
\affil[3]{Department of Computer Science, University of California Los Angeles}
\affil[ ]{\tt \{sarik, xuezhema, galstyan\}@isi.edu  \tt zixiliu@usc.edu  \protect\\ \tt tuhin.chakr@cs.columbia.edu  \tt violetpeng@cs.ucla.edu}

\begin{document}
\maketitle
\begin{abstract}
Having engaging and informative conversations with users is the utmost goal for open-domain conversational systems.
Recent advances in transformer-based language models and their applications to dialogue systems have succeeded to generate fluent and human-like responses. However, they still lack control over the generation process towards producing contentful responses and achieving engaging conversations.  To achieve this goal, we present \textbf{DiSCoL} (\textbf{Di}alogue \textbf{S}ystems through \textbf{Co}versational \textbf{L}ine guided response generation). DiSCoL is an open-domain dialogue system that leverages conversational lines (briefly \textbf{convlines}) as controllable and informative content-planning elements to guide the generation model produce engaging and informative responses. Two primary modules in DiSCoL's pipeline are conditional generators trained for 1) predicting relevant and informative convlines for dialogue contexts and 2) generating high-quality responses conditioned on the predicted convlines.  
Users can also change the returned convlines to \textit{control} the direction of the conversations towards topics that are more interesting for them.
Through automatic and human evaluations, we demonstrate the efficiency of the convlines in producing engaging conversations.      
 
\end{abstract}  

\section{Introduction} \label{intro}
Over the past decade, users have actively engaged with dialogue systems to fulfill a wide range of requirements. On one hand, \textit{task-oriented dialogue systems} have assisted users in accomplishing specific tasks such as finding apartments and restaurants or even booking movie tickets \cite{gustafson2000adapt, gruenstein2007releasing, li2017end}. On the other hand, \textit{open-domain dialogue systems} have been extensively leveraged for psychotherapy counseling, entertainment, and even teaching foreign languages to users  \cite{zhou2020design, oh2017chatbot, sarosa2020developing}. In this work, we focus on the second group. 

\begin{figure}[t]
\centering
\includegraphics[width=\linewidth]{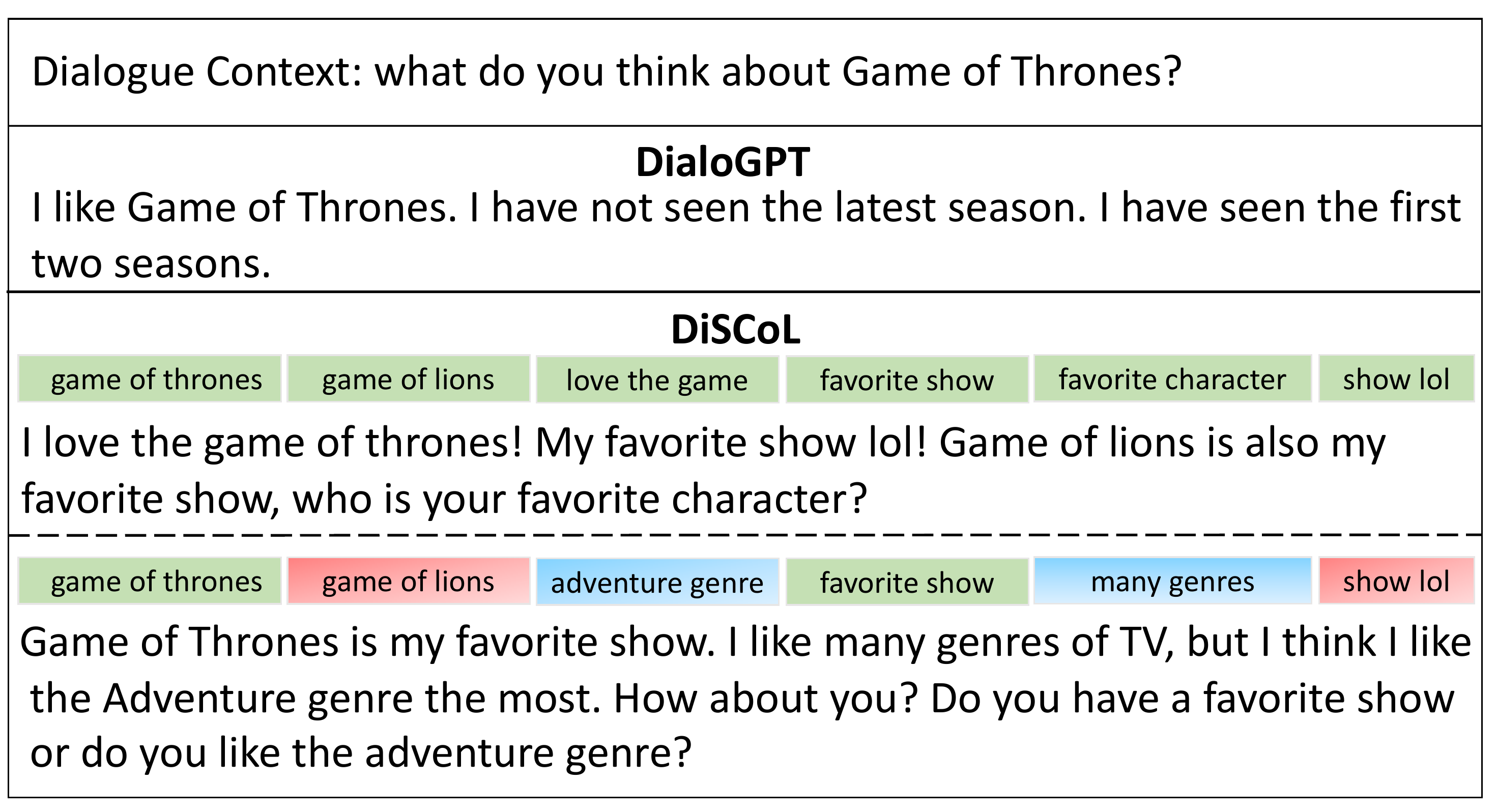}
\caption{A dialogue context and its three responses generated based on DialoGPT and our proposed DiSCoL system using originally inferred and manipulated convlines, respectively.
DiSCoL leverages convlines (depicted in colored boxes) to guide the generation model to encapsulate those informative contents. Our demo enables the user to edit or remove the inferred convlines (shown in blue for edits and red for removal) to guide the conversation towards its desired directions.}
\label{mot_img}
\vspace{-1.5em}
\end{figure}

\begin{figure*}[t]
\centering
\includegraphics[width=\linewidth]{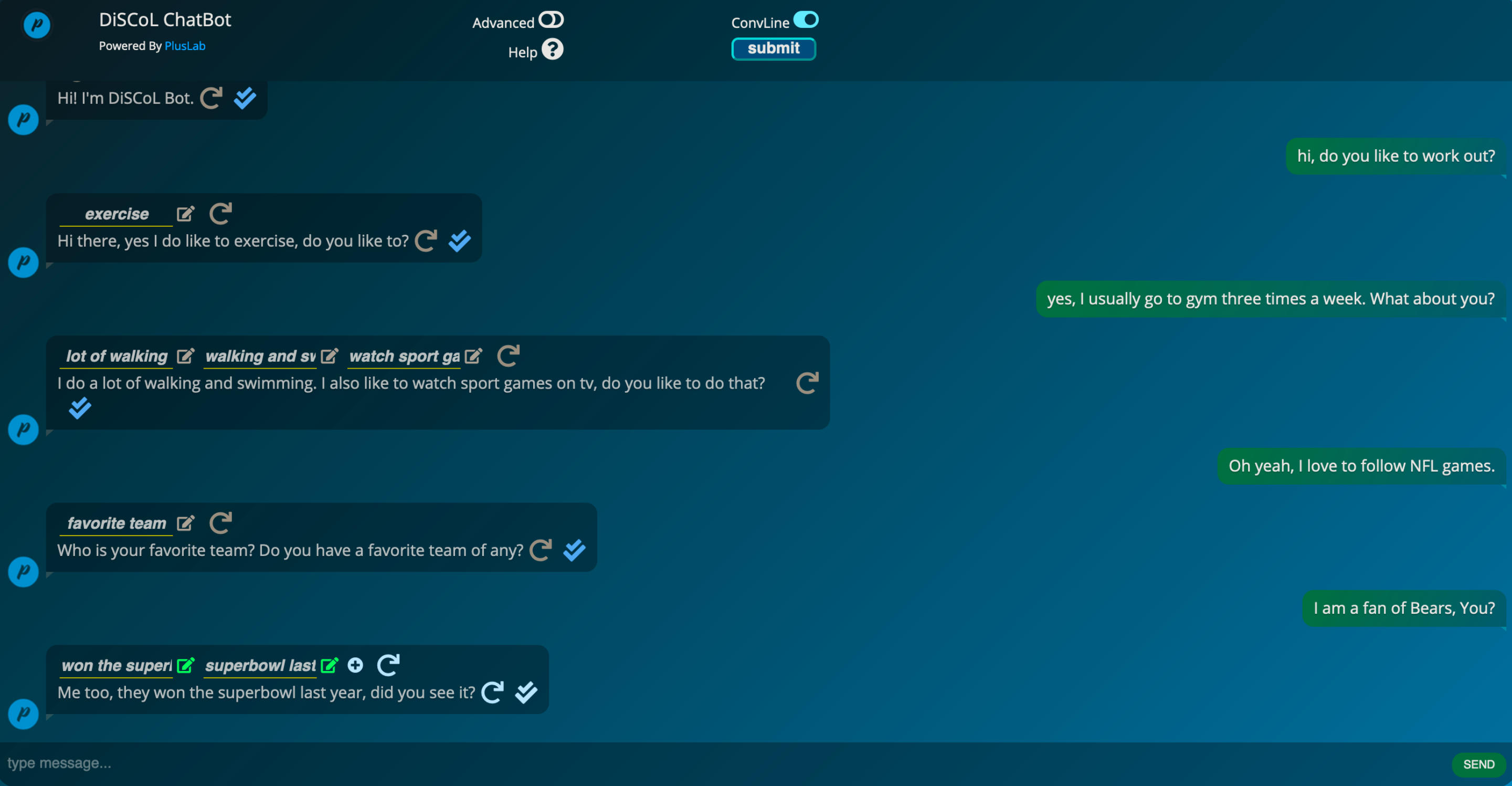}
\caption{A snapshot of the proposed DiSCoL system}
\label{sys_snap}
\vspace{-1.5em}
\end{figure*}
In the context of open-domain dialogue systems, neural-network-based generative models have outperformed retrieval-based systems by generating diverse and novel responses.
More recently large-scale language models with transformer-based architectures, such as GPT-2 \cite{radford2019language} and BART \cite{lewis2019bart}, have advanced the state of the art in Natural Language Generation and Dialogue Systems. Such models can be further enhanced by fine-tuning them on task-specific data, as it is the case of DialoGPT  
(dialogue generative pre-trained transformer) \cite{zhang2019dialogpt},
a neural conversational response generation model, trained on 147M conversation-like exchanges extracted from Reddit. Although generated responses by such models are fluent and locally coherent, usually they suffer from content poverty and often include less informative contents which make them not as interesting and engaging for the users. The first block in Figure \ref{mot_img} depicts an example of a generated response by DialoGPT. Indeed, these models do not provide users the possibility to have controls on the generation contents and guide the conversation towards users' desired direction.   

To alleviate this issue of generating informative and controllable responses, we propose DiSCoL that is an open-domain dialogue system with the intervention of \textbf{convlines} as primary elements to add \textit{control} for generating informative and content-rich responses. Convlines are abstract representations of utterances in the dialogues that can be used as content planning elements to form high-level contents of an utterance and guide the generator to incorporate these informative units in the generation (See colored boxes in Figure \ref{mot_img}). 
Content planning has also been beneficial in story generation task. These abstract representations known as storylines or story plots have been successful to guide the
language models produce more coherent and fluent stories \citep{yao2019plan,goldfarb2019plan,fan2019strategies, goldfarb2020content,rashkin-etal-2020-plotmachines}.

DiSCoL is
composed of four main neural-network-based modules (See Figure \ref{sys_arch1}). The first two modules are designed to extract entities and topics of the dialogue context. The third module is a fine-tuned conditional generator that learns to take the dialogue context and previously extracted information and predict convlines that would be leveraged in the response generator module. Similar to convlines generator, response generator is a conditional auto-regressive language model that generates response conditioned on the dialogue context and its convlines, entities, and topics extracted from previous modules. The middle block of Figure \ref{mot_img} exhibits the generated response for the inferred convlines shown in green boxes. In the interactive setting of our devised demo from which a snapshot is shown in Figure \ref{sys_snap}, we provide the facility that the user can manipulate the predicted convlines to direct the conversation towards its topics of interest. The last block in Figure \ref{mot_img} depicts the removed and edited convlines (red and blue boxes) that led the generator to generate a slightly different response by taking into account the applied adjustments. 

We validate DiSCoL on the Topical chat dataset \citep{gopalakrishnan2019topical} using 
 both human and automatic evaluations. Our results demonstrate the superiority of DiSCoL over DialoGPT in terms of generating higher quality responses, thus indicating the usefulness  of convlines as dialogue control mechanisms for generating more engaging responses. We release the source code and trained models to facilitate the future dialogue research. \footnote{Github Link: \url{https://github.com/PlusLabNLP/Dialogue_System_Hackathon}}

\section{System Architecture} \label{sys_arch}

\begin{figure*}[t]
\centering
\includegraphics[width=\linewidth]{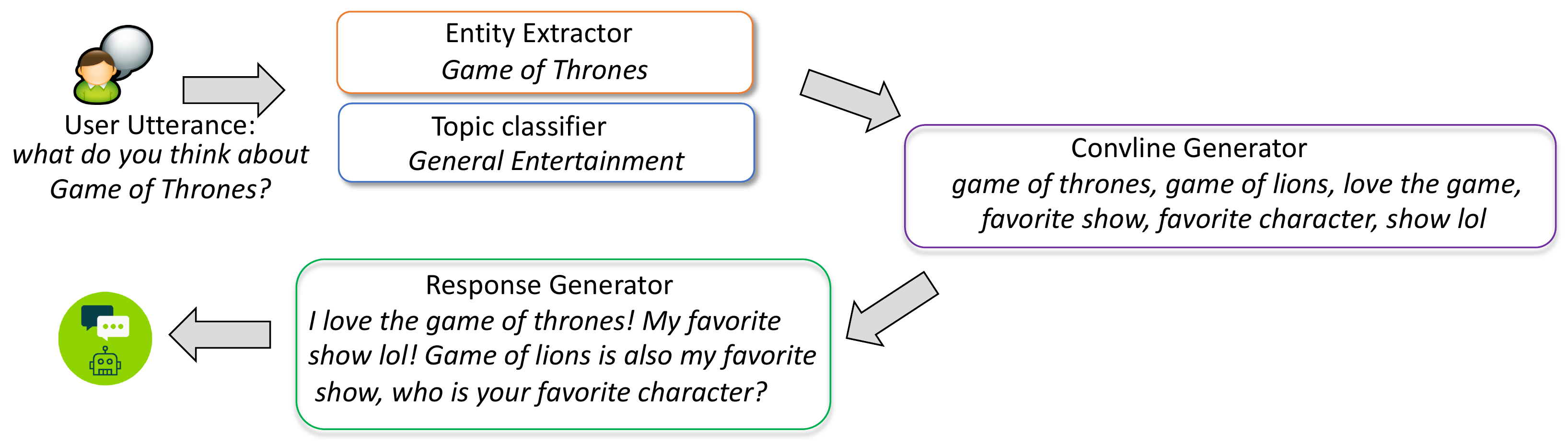}
\caption{Architecture of DiSCoL system}
\label{sys_arch1}
\vspace{-1.5em}
\end{figure*}

The architecture of our proposed DiSCoL demo system and its modules are depicted in Figure \ref{sys_arch1}. 
A user converses with the system by writing an utterance as an input. This utterance passes through all the modules and in each module some new information such as its extracted entities, topics, and convlines are augmented. The last module, response generator, incorporates all this information to generate a response as the output of the system. In this section, we explain each module in detail. 

\subsection{Entity Extractor}
One of the principal components in the conversational systems is the set of entities that both interlocutors are interested to converse about. It is crucial that the system can identify the main entities from the dialogue context and try to continue the conversation by providing more relevant information or even expressing its opinions and impressions regarding them. Therefore, in DiSCoL we take the user's utterance as the dialogue context and extract its entities. This task in known as a named entity recognition (NER) task, where each token in the text is classified into one of the predefined classes such as a person, organization, location or other.

Toward this goal, we leverage the BERT model \citep{devlin2018bert} fine-tuned on CoNLL-2003 dataset \citep{sang2003introduction}, which is a well-known corpus for NER task.\footnote{We leverage fine-tuned BERT model provided by Huggingface (https://github.com/huggingface/transformers).} We detokenize the output of the fine-tuned BERT model to get the original version of entities' tokens and disregard the predefined classes of entities since in our case they do not augment additional benefits. As it is shown in Figure \ref{sys_arch1}, all entities with labels other than \textit{O} are returned from the entity extractor module.

\subsection{Topic Classifier}
Knowing the topic that the user is enthusiastic to discuss would be beneficial for the dialogue system to generate utterances about that specific topic. The blue box in Figure \ref{sys_arch1} represents the topic classifier that takes the user's utterances and predicts the most relevant topics from a predefined set. These topics are later used for predicting convlines and consequently generating responses.

Due to the proven effectiveness of the BERT model \citep{devlin2018bert} and its wide applicability in many classification tasks, we incorporate it into the topic classifier module of DiSCoL. We fine-tune 
BERT model on pairs of utterances and their aligned topics 
with the main goal of minimizing the cross-entropy loss.

\subsection{Convline Generator}
DiSCoL's main contribution is in the convline generator module that is depicted as the purple box in Figure \ref{sys_arch1}. Convlines are abstract representations or content plans of utterances throughout the conversation. These representations that are known as storylines or story plots in the story generation context have recently posited their efficiency in generating higher quality stories \citep{yao2019plan,fan2019strategies, goldfarb2020content,rashkin-etal-2020-plotmachines}. Story generation models leverage plan-and-write framework that is successful in generating fluent and informative stories by the intervention of storylines as an intermediate step. 
In this work, we follow the same idea but in the context of conversational systems. In particular, we aim to show that the controlled generation of high-quality utterances by planning in advance and leveraging useful abstract-level convlines can be beneficial for dialogue systems as well.

To compose the convlines as the main component in the convline generator module, we extract sequences of important words in each utterance from existing human-human conversational data. We pursue the YAKE \citep{campos2018yake} method that relies on the text's statistical features to extract the most important keywords of an utterance. It has shown its superiority versus other state-of-the-art unsupervised approaches such as TF-IDF and RAKE \citep{rose2010automatic}. 

In order to train the convline generator, we extract pairs of ($u_i, r_i$) as a set of consecutive pairs of dialogue context utterances and their corresponding ground-truth responses in the human-human conversational data. For each dialogue context utterance ($u_i$), we extract its entities ($e_i$) and topics ($t_i$) using the entity extractor and topic classifier modules. Each response ($r_i$) is replaced by its convlines ($c_i$) obtained by the YAKE algorithm. The constructed input data are in ($u_i, e_i, t_i, c_i$) format. 

\begin{figure*}[t]
\centering
\includegraphics[width=\linewidth]{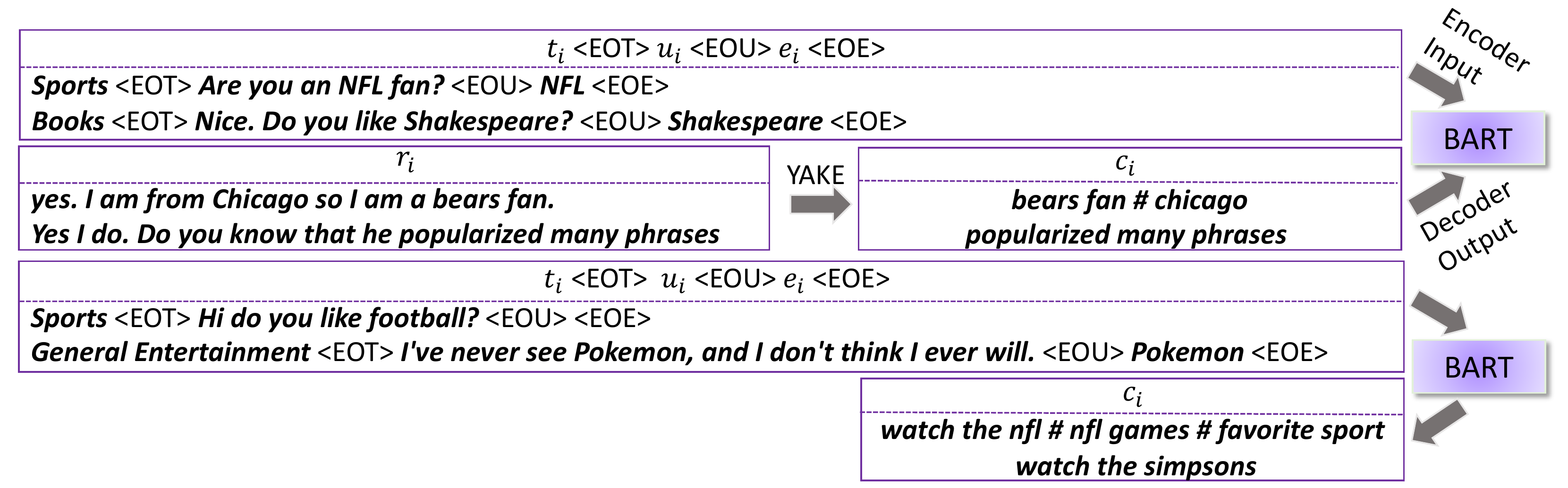}
\caption{Architecture of the convline generator during training and inference time}
\label{conv_line}
\vspace{-1.5em}
\end{figure*}

The convline generator is a conditional model that generates the most probable convlines given the provided dialogue context utterance and its entities and topics. To this end, we apply BART \citep{lewis2019bart} a state-of-the-art pre-trained sequence-to-sequence generative model. It combines a bidirectional encoder as that of BERT \cite{devlin2018bert} to encode the input and a GPT like \citep{radford2018improving} auto-regressive decoder model to generate convlines as the output. The top block in Figure \ref{conv_line} encapsulates the training process of the convlines module. We fine-tune BART on the constructed training data with the objective of minimizing the negative log likelihood that is shown in equation \ref{eq:loss_conv}.
\begin{equation}\label{eq:loss_conv}
L_{line\_gen} = -log \sum_{i = 1}^{n} P(c_i|u_i, t_i, e_i) \end{equation}
During inference, the fine-tuned BART model takes the user's utterance plus its inferred entities and topics to predict the most probable convlines as it is depicted in the bottom block of Figure \ref{conv_line}. We use top-k sampling \cite{fan2019strategies} with $k=5$ and a temperature of $0.7$ for the generation.

\subsection{Response Generator}
The last module in DiSCoL system's pipeline is the response generator that is identical to convline generator except for the type of inputs and outputs. 
The response generator takes the dialogue context utterance, its convlines and topics as inputs and generates response conditioned on those data.
\begin{equation}\label{eq:loss_gen}
L_{resp\_gen} = -log \sum_{i = 1}^{n} P(r_i|u_i, t_i, c_i) 
\end{equation}
During training, we provide utterances, their topics and convlines extracted from YAKE to the BART model and fine-tune this pre-trained conditional generator. As it is shown in equation \ref{eq:loss_gen}, the training objective is to maximize the probability of generating ground-truth responses given their context utterances, topics followed by convlines. 

During inference, the generator attempts to produce the most probable responses that include convlines returned by the convline generator module. 

\section{System Implementation} \label{exp_setup}
We test our system on Topical-Chat dataset \citep{gopalakrishnan2019topical} that includes knowledge-grounded human-human conversations covering a set of 8 different topics. This dataset has been collected by employing Amazon Mechanical Turk (AMT) workers who have been provided with specific entities and some external knowledge (Wikipedia lead sections, Washington Post articles, or some Reddit fun facts) to chat about. Therefore, each utterance in the conversation is either based on provided knowledge sources or the user's personal knowledge. Overall, 261 popular entities spanning 8 various topics (Fashion, Sports, Books, Politics, and etc.) have been selected for the dataset collection. We add \textit{General} topic for utterances (e.g. greetings) that do not include any specific contents such as \textit{"hi, how are you today?"}.

\subsection{Topic Classification Data} 
Although each utterance in the Topical-Chat dataset comes from either external or personal knowledges, it lacks specified topics. These topics are necessary for DiSCoL modules. We manually match all 261 entities in the external knowledges to one of the topics in the predefined set and easily label all utterances about entities of those external knowledges to their matched topics. This results in about 78\% of overall 188,378 (\textit{easy\_set)} utterances to be easily matched with topics. 
As an example, the utterance \textit{"Do you know Tom Brady"} is about \textit{"Tom Brady"} entity that is an indication of the \textit{"Sports"} topic. The remaining challenging utterances are based on personal knowledges that their entities are not directly specified. We pursue the following context-based heuristics to label such \textit{challenging\_set} utterances with their relevant topics. If the utterance's neighbors are from \textit{easy\_set} and share the same entity, we assign that entity's topic to the utterance, while in the case of neighbors containing different entities, we label the given utterance with both utterances' topics. If the previous rules do not apply to an utterance in the \textit{challenging\_set}, we use the most frequent topic in the dialog as its topic.
\begin{table}[t]
\centering
\small
\begin{tabular}{cccc}
\hline
\textbf{Uttr.}  & \textbf{Easy\_set} & \textbf{Challenging\_set} & \textbf{General\_uttr.} \\ \hline
188,378 & 146,370 & 5,323 & 5,966 \\
\hline
\end{tabular}
\caption{Statistics of different groups of utterances (uttrs.) in Topical chat dataset}
\label{topic_stats}
\vspace{-1em}
\end{table}
    
    
In parallel to the above heuristics and in order to improve the quality of assigned topics, we apply a keyword-based classifier that classifies \textit{challenging\_set} utterances with appropriate topics. The keyword-based classifier retrieves the most similar entity from the overall 261 entities to each utterance's keywords using their BERT embeddings. The manually matched topics for the retrieved entity are assigned to the utterance. We only consider 5323 \textit{challenging\_set} utterances that their adapted labels based on both context-based heuristics and keyword-based classifier are the same (See statistics in Table \ref{topic_stats}). We fine-tune the BERT model as the topic classifier for 10 epochs and get an accuracy of 85.55 on the validation set.

\subsection{Convline Generator Data}
Convlines are necessary components in the training of the DiSCoL system. We leverage YAKE \citep{campos2018yake} for retrieving discourse keywords representing convlines. YAKE assigns importance score to tokens in a text by following an unsupervised approach that builds upon features extracted from the text \citep{campos2018yake}. In this model, a set of features are computed for each term in the text. Subsequently, a list of candidates (n-grams of tokens) is created. Next, the Levenshtein distance is used to remove duplicate keywords. At the enc, the aggregation of tokens scores in each keyword represents the keyword's scores. Keywords with lower scores are returned as the text's salient convlines. Using YAKE we  
generate a contiguous sequence of 1, 2, and 3-grams candidate convlines.
We extract 3-grams convlines, followed by extracting 2-grams and 1-gram that are not included in the previously returned keywords. 
We fine-tune BART-large for both convlines and response generator models for 3 epochs and checkpoint the best epoch based on validation perplexity.\footnote{We fine-tune BART model using https://github.com/pytorch/fairseq}

\section{Experimental Results} \label{exp_results}
We compare the performance of DiSCoL system versus DialoGPT as one of the strongest recent baselines, that has shown its efficiency in generating consistent and relevant responses.

\begin{table}[t]
\centering
\small
\begin{tabular}{cccc}
\hline
\textbf{Dialogue Context}  & \textbf{Annotators} & \textbf{Kappa} & \textbf{Pearson} \\ \hline
100 & 33 & 0.44 & 0.5 \\
\hline
\end{tabular}
\caption{Statistics and inter-annotator agreements of AMT evaluations on DiSCoL and DialoGPT performances.}
\label{amt_stats}
\vspace{-1em}
\end{table}

\subsection{Metrics} 
To explore the efficiency of our proposed controlled response generation, we apply both automatic and human evaluations.
\begin{figure}[t]
\centering
\includegraphics[width=\linewidth]{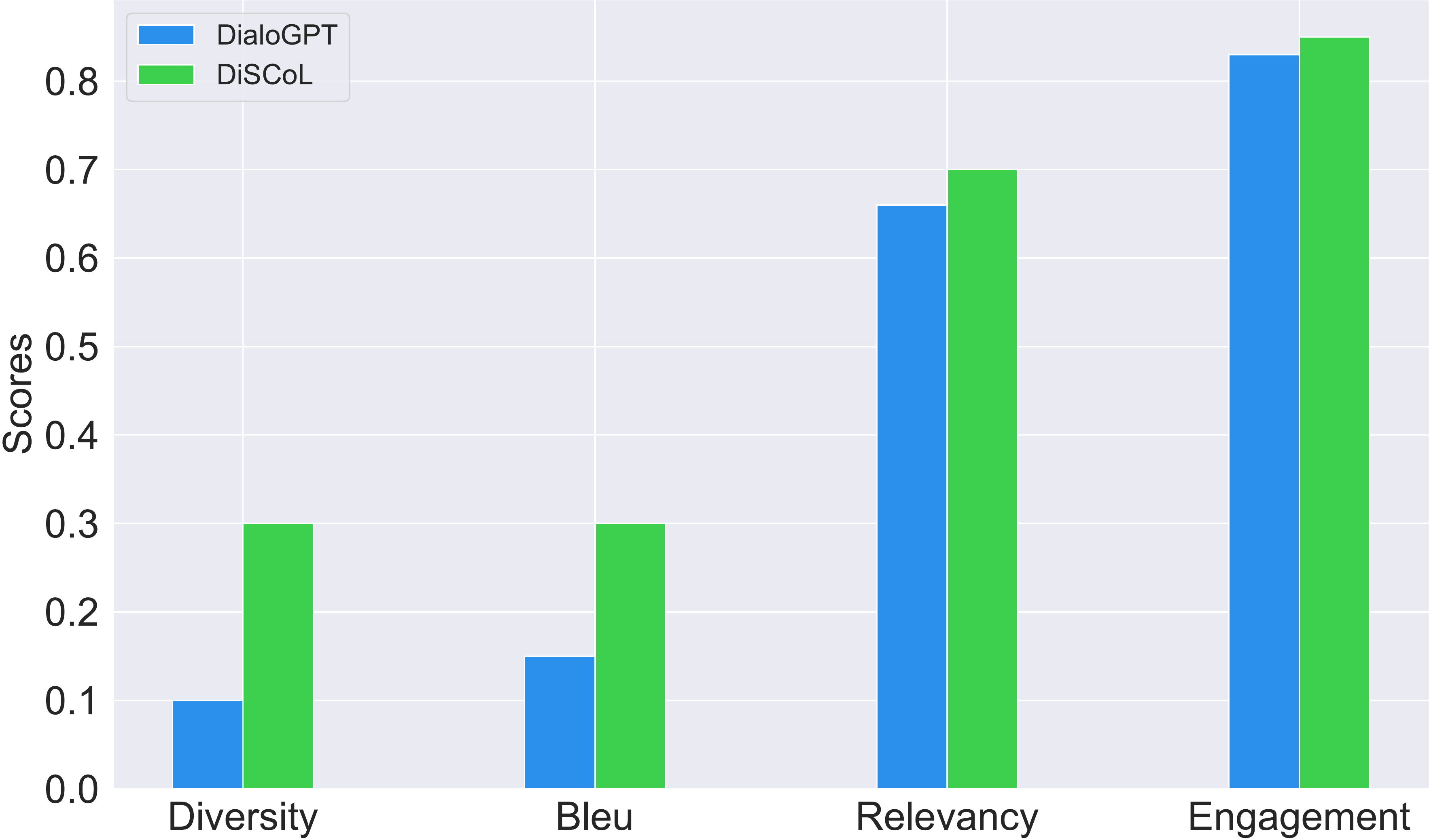}
\caption{ Automatic evaluations on responses generated by DiSCoL and DialoGPT systems}
\label{auto_eval}
\vspace{-1.5em}
\end{figure} 
\subsection{Automatic Evaluations}
Due to the multi-faceted nature of dialogue quality, it is necessary to do the evaluation from different aspects \cite{see2019makes, mehri2020unsupervised}. To this end, we compare the quality of DiSCoL and DialoGPT generated responses through computing different metrics. We conduct automatic evaluations and compute evaluation metrics on 23,530 consecutive utterance pairs (dialogue context utterances and their ground-truth responses) of the Topical chat test set. The measured metrics are averaged over all utterance pairs within the test set.  We compute BLEU-3 \cite{papineni2002bleu} to evaluate the similarity of generated responses to ground-truth responses based on the 3-grams overlaps. Due to the one-to-many essence of open-domain dialogue systems and the imperfection of such word-overlap metrics \cite{liu2016not, ghazarian2019better, mehri2020unsupervised}, we also focus on three main aspects: diversity, relevancy, and engagingness as better indications of systems performances. 

Diversity measures the percentage of distinct generated tokens by each model. \citet{li2015diversity} proposed distinct-2 that computes distinct bi-grams divided by the total number of generated words. Relevancy utilizes both dialogue context utterance and the generated response to deliberate how much it is relevant to the given utterance \cite{tao2018ruber, ghazarian2019better}. 
We use the contextualized Ruber metric for this purpose \cite{ghazarian2019better}. At the end, since in open-domain dialogue systems, it is necessary to have both relevant and interesting responses to make the user feel satisfied \cite{ghazarian2020predictive}, we further validate systems based on the engagingness of responses.  We compute engagingness as the probability score of the engaging class predicted by \citet{ghazarian2020predictive}'s proposed engagement classifier.   
 
\begin{figure}[t]
\centering
\includegraphics[width=\linewidth]{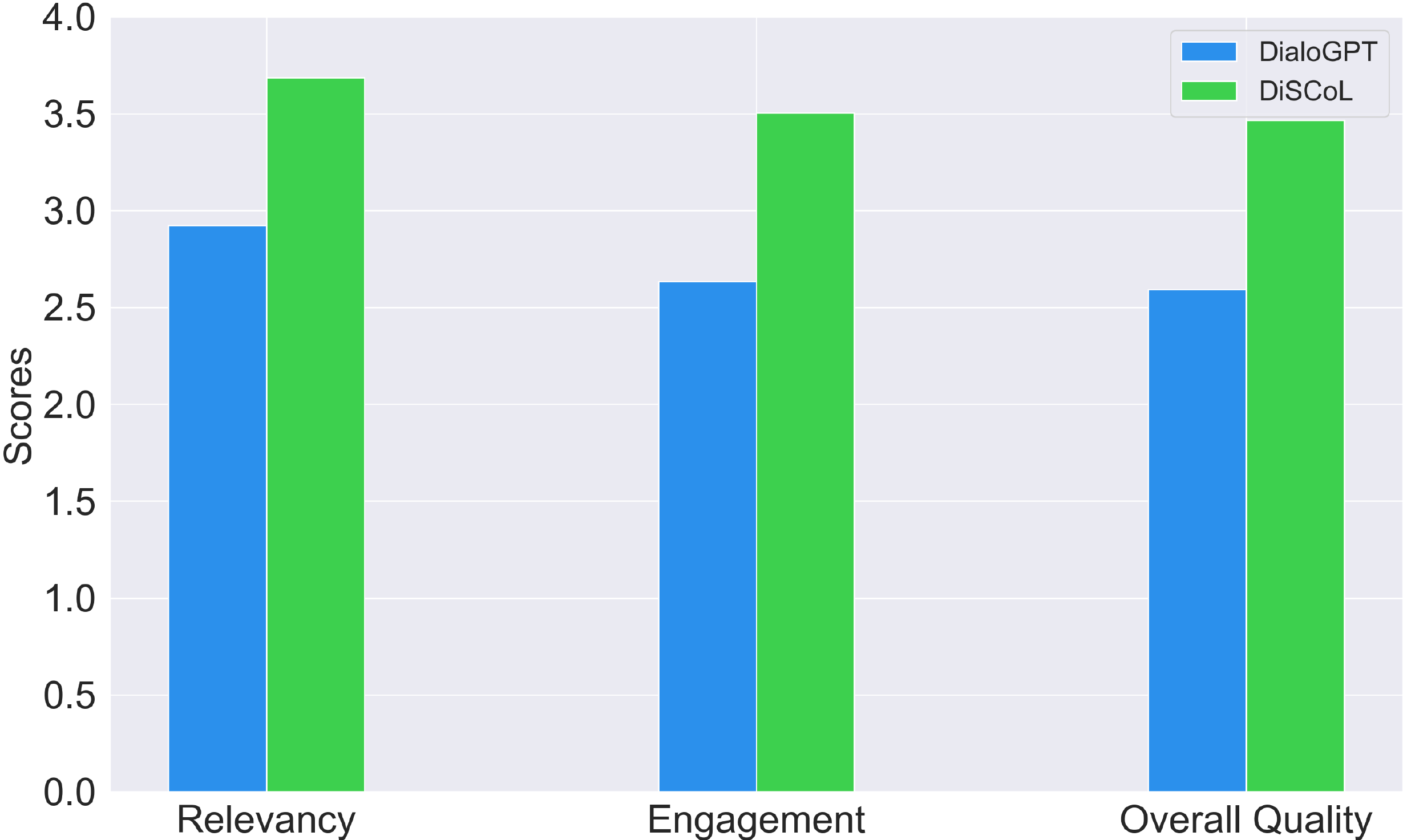}
\caption{Human evaluations on responses generated by DiSCoL and DialoGPT systems }
\label{human_eval}
\vspace{-1.5em}
\end{figure} 
\subsection{Human Evaluations}
We extend our evaluations by running AMT experiments to report human judgments on systems' qualities. We randomly select 100 dialogue context utterances from the Topical chat test set. For each given dialogue context utterance, we ask three AMT workers to rate DiSCoL and DialoGPT's generated responses by keeping these systems anonymous. Participants rate the relevancy, engagingness, and overall quality of each response on a 5-point Likert scale (1 
indicating irrelevant/not engaging and low-quality response). The statistics of the AMT experiment is shown in Table \ref{amt_stats}. 

\subsection{Results}
\paragraph{Automatic Evaluation.} Figure \ref{auto_eval} depicts the average scores of diversity, BLEU, relevancy, and engagingness resulted from automatic evaluation metrics for all the generated responses of DiSCoL and DialoGPT systems. 
The strength of DiSCoL is noticeable from its higher BLEU score and 
more diverse, relevant, and engaging responses. Overall, the diversity is low due to the limited distinct topics considered in the Topical chat dataset. 
The BLEU metric is low for both systems which shows its inadequacy in the open-domain evaluations; where a response can be super appropriate and at the same time not similar to the ground-truth response.

\paragraph{Human Evaluation.} The bars in Figure \ref{human_eval} demonstrate the average of human annotations for different qualities of generated utterances. Each response's score is the mean aggregation of three annotators' ratings. According to Figure \ref{human_eval}, annotators appraise responses generated by DiSCoL with higher scores in terms of relevancy, engagingness, and overall quality. This could be an evidence for the positive impact of incorporating convlines to guide the dialogue system towards generating controllable, relevant, and contentful responses that infuse the user to converse for a longer time.

\section{Conclusion} \label{conclusion}
We have introduced DiSCoL that leverages convline as an intermediate step towards generating more informative and controllable responses in dialogues. These convlines are predicted and subsequently leveraged in the response generation process. Additionally, DiSCoL allows users to 
manipulate convlines towards their favorite conversational direction. Our findings show that in contrast to other transformer-based dialogue systems that do not have content plannings, our system takes the advantage of such a principled structure to have better and more engaging conversations with users. 
\section{Ethics} \label{ethics}
Through the entire phases of the conducted research and developed DiSCoL system, all co-authors were agreed and adhered to \textit{ACM Code of Ethics}. Our effort was to ensure we stuck to the conscience of the profession and considered the Code principles. We certify that this system and all the presented evaluations are compatible with the provided code.

\paragraph{DiSCoL System's Development} 
The main contribution of our proposed DiSCoL system is to augment controllable response generation with the intervention of convlines that leads the generation towards producing more relevant and interesting responses. Indeed, DiSCoL provides an opportunity for users to manipulate the convlines and guide the system to continue the conversation in the user's favorite direction. All DiSCoL's modules leverage pre-trained large language models such as BART \cite{lewis2019bart} and fine-tune them on recently proposed Topical chat dataset \cite{gopalakrishnan2019topical}. One potential harm that DiSCoL could cause is its feasibility to generate improper responses conditioned on the inferred convlines with abusive contents. Since the convline and response generators are BART models finetuned on human-human conversations that do not encompass profanity and inappropriate content (\cite{gopalakrishnan2019topical}), hence the convlines that indeed are important informative units of the utterances would be free of bias and obscene content. However, there still is a possibility of dual-usage attacks by augmenting conversations with offensive languages to fine-tune the generators and teach them to generate such inappropriate content. The identification of such attacks that could occur in almost all learnable models and the way to overcome them by itself is a distinct and huge research area that is out of this paper's scope.

\paragraph{DiSCoL System's Evaluation} Alongside the automatic evaluation for demonstrating the efficiency of controllable generations using convlines, we further collected human annotations by conducting Amazon Mechanical Turk (AMT) experiments. We provided different systems responses for given utterances while keeping systems anonymous and asked users to rate responses by considering different aspects that had been explained in the AMT surveys. We estimated the average time users would spend on each survey and fairly compensated them according to the hourly wage. 

We kept the privacy of all AMT turkers who participated in the experiments. Our experiments did not have the requisite to know the user's personal information, therefore their personal information including their genre, ethnicity, and etc. are not revealed. This fades the necessity for IRB approvals. 

Our system's target is NLP open-domain conversational AI community. Its main goal is to achieve engaging conversations with the incorporation of convlines and increase the user's ability to control the generation process. Likewise other proposed dialogue systems, we anticipate specific failure modes specifically for novel conversations on new topics. Lifelong learning in dialogue systems which is not the focus of this work is a research area that attempts to enhance conversation systems' ability to deal with such novel scenarios.

\bibliography{anthology,custom}
\bibliographystyle{acl_natbib}

\appendix


\typeout{get arXiv to do 4 passes: Label(s) may have changed. Rerun}
\end{document}